\def\year{2019}\relax
\documentclass[letterpaper]{article} 
\usepackage{aaai19}  
\usepackage{times}  
\usepackage{helvet}  
\usepackage{courier}  
\usepackage{url}  
\usepackage{graphicx}  

\usepackage{booktabs}
\usepackage{multirow}
\usepackage{float}
\usepackage[ruled]{algorithm2e}
\usepackage{amsmath}
\usepackage[skip=0pt]{caption}	
\usepackage{algorithmic}
\usepackage{subcaption}
\usepackage{enumitem}
\usepackage{array}
\usepackage[final]{animate}

\begin{document}
%
\title{Mode Variational LSTM Robust to Unseen Modes of Variation: Application to Facial Expression Recognition}
\author{Wissam J. Baddar\ and  Yong Man Ro \thanks{Corresponding author: Yong Man Ro \{ymro@kaist.ac.kr\} .} \\
Image and Video Systems Lab., Electrical Engineering, \\
KAIST, South Korea\\
\{wisam.baddar,ymro\}@kaist.ac.kr\\
}

\maketitle

\begin{abstract}
	Spatio-temporal feature encoding is essential for encoding the dynamics in video sequences. Recurrent neural networks, particularly long short-term memory (LSTM) units, have been popular as an efficient tool for encoding spatio-temporal features in sequences. In this work, we investigate the effect of mode variations on the encoded spatio-temporal features using LSTMs. We show that the LSTM retains information related to the mode variation in the sequence, which is irrelevant to the task at hand (e.g. classification facial expressions). Actually, the LSTM forget mechanism is not robust enough to mode variations and preserves information that could negatively affect the encoded spatio-temporal features. We propose the mode variational LSTM to encode spatio-temporal features robust to unseen modes of variation. The mode variational LSTM modifies the original LSTM structure by adding an additional cell state that focuses on encoding the mode variation in the input sequence. To efficiently regulate what features should be stored in the additional cell state, additional gating functionality is also introduced. The effectiveness of the proposed mode variational LSTM is verified using the facial expression recognition task. Comparative experiments on publicly available datasets verified that the proposed mode variational LSTM outperforms existing methods. Moreover, a new dynamic facial expression dataset with different modes of variation, including various modes like pose and illumination variations, was collected to comprehensively evaluate the proposed mode variational LSTM. Experimental results verified that the proposed mode variational LSTM encodes spatio-temporal features robust to unseen modes of variation.  
\end{abstract}

\section{Introduction}
	One of the key features of deep learning is trying to learn the latent features from sample (training) data. However, encoding features that represent all types of variation that could occur in a data sample is hard to achieve \cite{1,2}. For example, when considering a population of face images, the face images would have different identities, expressions, poses and illumination variations. The resulting face image could be considered as a multifactor confluence of all those modes of variation \cite{3}.

Statistical methods have tried to provide a single mode of variation. Most popular of those statistical methods are: factor analysis \cite{4}, principal component analysis (PCA) \cite{5} and singular value decomposition (SVD) \cite{6}. However, most forms of visual data (images and videos) have many different, and possibly independent, modes of variation. This makes it difficult for single mode of variation methods (such as PCA) to be able to represent all the variations in such visual data \cite{3}.

To reduce the negative effect of modes of variation, several supervised methods have been proposed to disentangle independent modes of variation and extract them from visual data \cite{7,8,9,10,11}. Such supervised methods disentangle multiple multilinear (tensor) decompositions of the data to represent all the variations \cite{3}. The high order SVD (HOSVD) \cite{8}  was proposed to identify different modes of variation in face images, by decomposing carefully designed data tensors (data tensors for identities, data tensors for expressions and data tensors for poses). This method is known as TensorFaces \cite{12}. The main drawbacks of the aforementioned supervised methods are: (1) they require labels of the modes of variation available at the training time. (2) They require the same number of samples  under all modes of variation (e.g., the same face under different expressions, poses etc.). Therefore, their applicability is limited to well-organized data, usually captured in well-controlled conditions \cite{3}. Moreover, such methods were devised for disentangling modes of variation in only spatial features of images. 

Different from previous methods that were designed to suppress the negative effect of modes of variation in images, we investigate the effect modes of variation has on the spatio-temporal features encoded from the sequence dynamics via LSTM. We show our proposed method on the task of facial expression recognition (FER). To that end, we first show that, by continuously feeding a static sequence (obtained by replicating a frame in the test sequence which could have mode of a variation) into the LSTM, a tangible representation of the mode of variation could be obtained. Based on that observation, we modify the structure of the LSTM to include a static sequence path (representing the mode of variation) that encodes a bias induced by the modes of variation in the sequence. We call the LSTM structure including this bias as mode variational LSTM. The mode variational LSTM suppresses the effect of variation, and results in spatio-temporal features robust to unseen variations. The contributions of this paper are summarized as follows:

\begin{enumerate}
\item We investigate the effect of different modes of variation on the spatio-temporal features encoded via LSTM. We show that, despite the forget mechanism in the LSTM, the obtained spatio-temporal features suffer from a different biases based on the different mode of variation. Accordingly, a visualization of the effect of each of multiple modes of variation (illumination, pose and appearance variations) is provided.
\item To reduce the effect of the bias induced by modes of variation on the encoded spatio-temporal features, we devise the mode variational LSTM. The proposed mode variational LSTM includes a static sequence path that encodes the bias induced by the mode of variation in the input sequence in a separate cell state. The encoded bias is then suppressed by a shared output gate (shared between the dynamics sequence path and the static sequence path). As a result, the mode variational LSTM encodes spatio-temporal features robust to variations unseen during the training.
\item A rich dynamic facial expression dataset was collected and made publicly available called the KAIST face multi-pose multi-illumination (KAIST Face MPMI) dataset. The videos in the dataset were collected under different modes of variation (illumination, pose and appearance variations).
\end{enumerate}

\section{The Effect of Mode Variations on the Spatio-temporal Features Encoded via LSTM}
\begin{figure*}[!bt]
    \centering    
    \begin{subfigure}[b]{1.0\textwidth}
		\centering
		\includegraphics[width=0.8\textwidth,keepaspectratio]{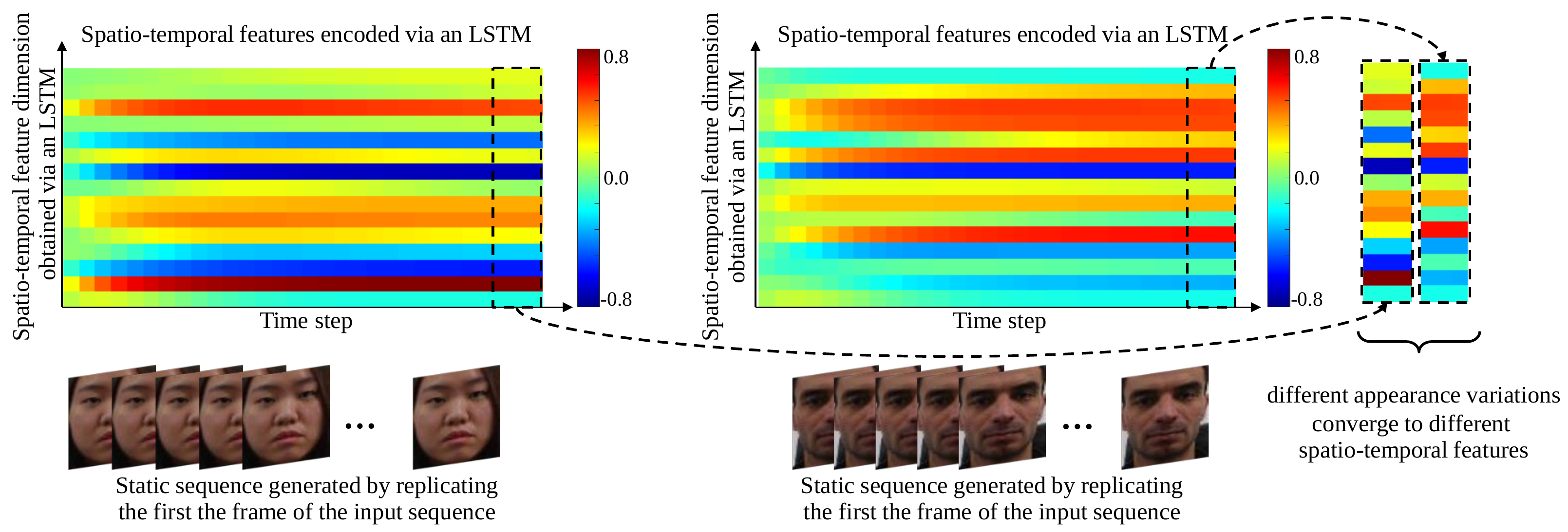}
        \caption{}
        \label{fig1:a}
	\end{subfigure}

    \begin{subfigure}[b]{1.0\textwidth}
        \centering
		\includegraphics[width=0.8\textwidth,keepaspectratio]{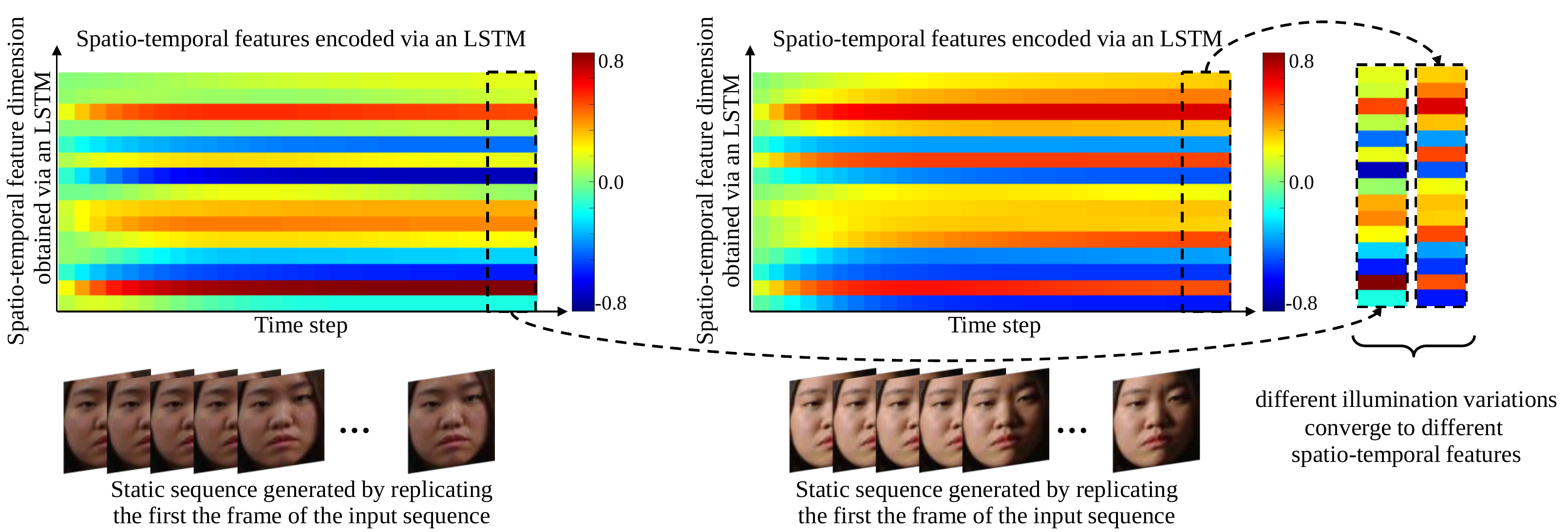}
        \caption{}
         \label{fig1:b}
	\end{subfigure}%

	\begin{subfigure}[b]{1.0\textwidth}
        \centering
		\includegraphics[width=0.8\textwidth,keepaspectratio]{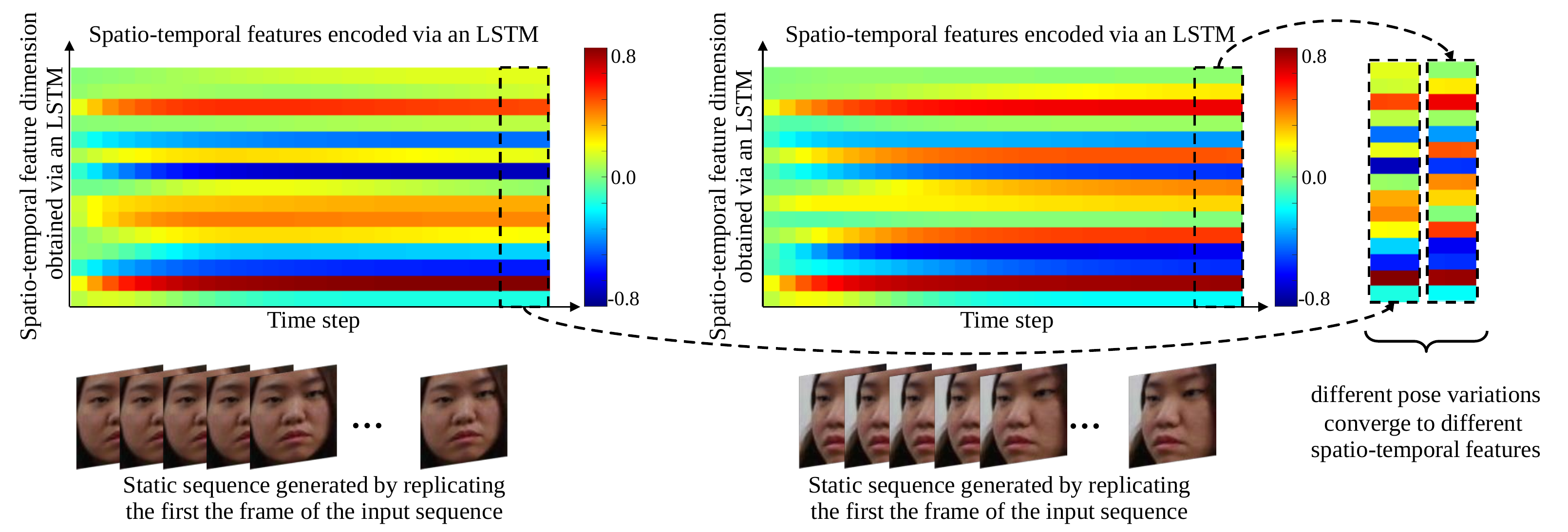}
        \caption{}
         \label{fig1:c}
	\end{subfigure}%
\caption{The effect of mode variations on the spatio-temporal features encoded via LSTM. The spatio-temporal features encoded from a pair of static sequences with (a) subjects appearance variation (b) illumination variation and (c) pose variation. In each feature graph, the first 15 spatio-temporal feature dimensions are shown. Each row represents one dimension of the 15 spatio-temporal feature dimensions. The static sequence was generated by replicating a frame 30 times. Best viewed in color.}
\label{fig:1}
\end{figure*}

	Before describing the proposed mode variational LSTM, we investigate the effect of mode variations on the spatio-temporal features encoded with LSTMs. To that end, we utilize an LSTM pre-trained for a classification task. In particular, we use a pre-trained LSTM to classify facial expressions in video sequences \cite{13}. Naturally, it is expected that the LSTM should ignore variations that could negatively affect the classification (FER) performance via the forget gate mechanism \cite{14}. Such mode variations include subject appearance variations, pose variations and illumination variations. In practice, we have observed that such mode variations leak into the spatio-temporal features encoded by the LSTM as a certain bias. This bias negatively affects the discriminability of the learned spatio-temporal features. To confirm that observation, a pre-trained LSTM model for FER \cite{13} was utilized. Instead of feeding the LSTM a dynamic facial expression sequence, we fed it a static sequence. The static sequence was obtained by replicating a frame $N$ times. The model was used to obtain and compare the spatio-temporal features of pairs of static sequences. Each pair of sequences contained a single mode variation (i.e., pose variation, illumination variation and a subject appearance variation). The spatio-temporal features of the static sequences were obtained and visualized in Figure~\ref{fig:1}. The spatio-temporal feature graphs in the figures show the first 15 dimensions out of 512 spatio-temporal feature dimensions encoded by the LSTM. 
 
Since the sequence is static, it is expected that the features obtained from the static sequence using the LSTM would be constant.  Moreover, it is expected that the spatio-temporal features of the pair of static sequences with different mode variations should be similar due to the LSTM forget mechanism. The reason behind this assumption is the fact that the mode of variation is irrelevant to the expression classification task. Hence the effect of the mode variation on the encoded spatio-temporal feature should be forgotten. As shown in Figure~\ref{fig:1}, the features change for a number of frames and then converge into a static state (until the warm up time is complete). This can be attributed to the LSTM forget gate mechanism, and the corresponding LSTM hidden cell state updates \cite{14}. More importantly, it can be observed that the spatio-temporal features of each pair converge into different values. This shows that the mode of variation induces a bias in the encoded spatio-temporal features, which negatively affects the discriminability of the spatio-temporal feature.

In this paper, we propose a modification on the LSTM structure, which adds a new cell state and the corresponding gating functionality. This modification to the LSTM is dedicated to encoding the bias induced by the mode of variation in the current sequence and suppress the effect of that bias. In the experiments detailed in section 4.3, we quantitatively show that unseen modes of variation negatively affect the prediction. However, the proposed mode variational LSTM encodes more robust features to unseen variations and improves the prediction performance.

\section{Proposed Mode Variational LSTM}
	\subsection{Mode variational LSTM structure}
	\begin{figure}[!ht]
    \centering
    
    \begin{subfigure}[b]{1\linewidth}
                 \centering
                 \includegraphics[width=0.9\linewidth,keepaspectratio]{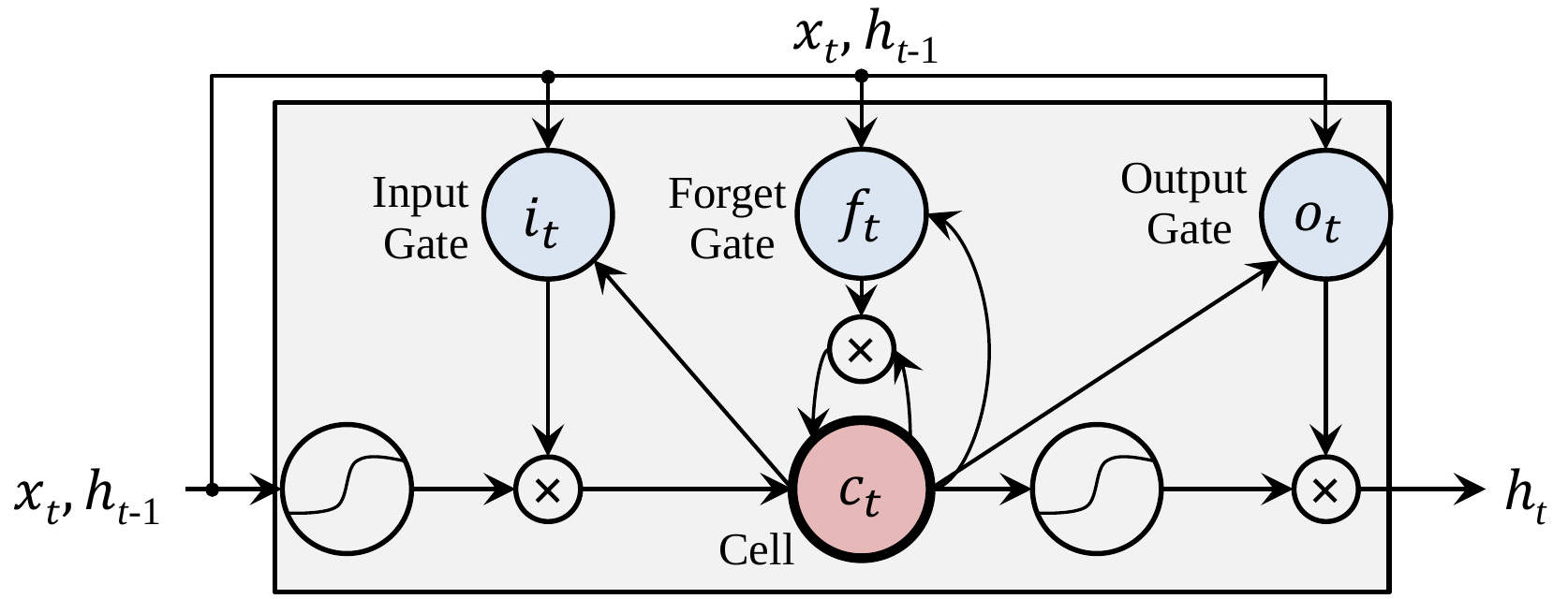}
                 \caption{}
             	 \label{fig2:a}
         \end{subfigure}
         

     \begin{subfigure}[b]{1\linewidth}
             \centering
             \includegraphics[width=0.9\textwidth,keepaspectratio]{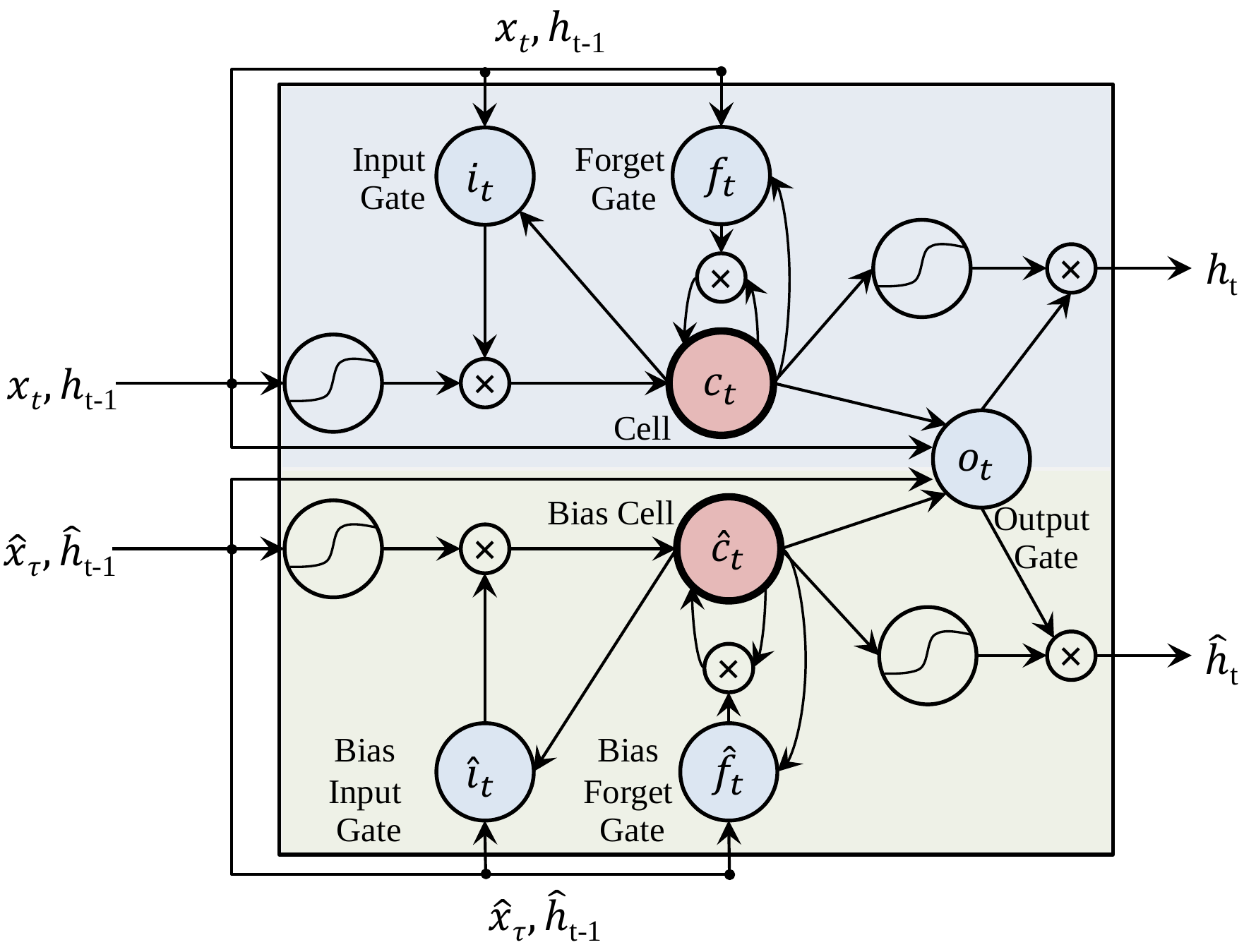}
             \caption{}
             \label{fig2:b}
     \end{subfigure}

	\begin{subfigure}[c]{1\linewidth}
             \centering
             \includegraphics[width=1\textwidth,keepaspectratio]{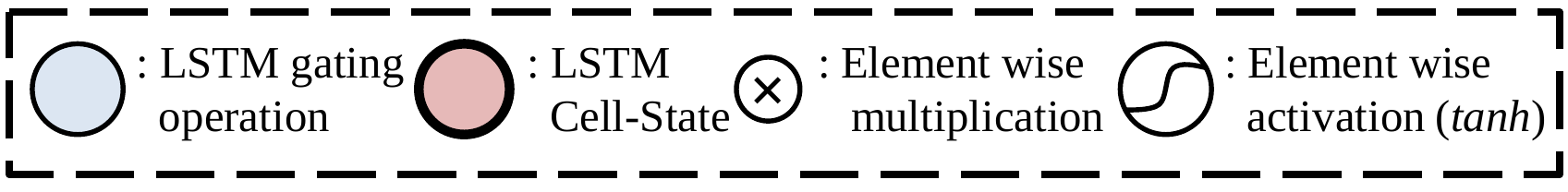}
             \label{fig2:c}
     \end{subfigure}
      
   \caption{Comparison between (a) the LSTM structure and (b) the proposed mode variational LSTM structure.}
\label{fig:2}
\end{figure}
	
	Figure~\ref{fig:2} shows a comparison between the LSTM defined in \cite{15} and the proposed mode variational LSTM. As shown in Figure~\ref{fig2:a}, the input gate determines which information should be added to the memory cell. The forget gate decides which information stored in the memory cell is important and should be retained (i.e., larger values are activated at the forget gate to retain information in its memory cell). As the network processes more frames (time steps), the memory cell state gradually absorbs the useful information related to the task in hand (e.g., FER). The output gate makes a latent feature representation of the output data related to the input frame at the current time step (e.g., a certain facial expression class). Note, that the LSTM shown in Figure~\ref{fig2:a} includes the peephole implementation as described in \cite{15} and is operated as follows:

\begin{equation} 
	\label{eq1}
	\begin{split}
		&i_t= \sigma{}(W_{xi}x_t+W_{hi}h_{t-1}+W_{ci}c_{t-1}+b_i),\\
		&f_t= \sigma{}(W_{xf}x_t+W_{hf}h_{t-1}+W_{cf}c_{t-1}+b_f),\\
		&c_t=f_tc_{t-1}+i_t tanh(W_{xc}x_t+W_{hc}h_{t-1}+b_c),\\
		&o_t= \sigma{}(W_{xo}x_t+W_{ho}h_{t-1}+W_{co}c_t+b_o),\\
		&h_t=o_t tanh(c_t),\\
	\end{split}
\end{equation}

\noindent where $i_t,f_t,c_t,o_t,h_t$ are the input gate, forget gate, cell state, output gate and the latent features at time $t$, respectively. $W_*$ and $b_*$ are the trainable weights and biases of the LSTM and $x_t$ is the input frame at time $t$. Finally, $\sigma(.)$ and $tanh(.)$ are the sigmoid and the hyperbolic tangent activation functions.

As shown in section 2, modes of variation can induce a certain bias that negatively affects the discriminability of the spatio-temporal features encoded by the LSTM. To improve the LSTM robustness to unseen variations, we propose adding a new memory cell $(\widehat{c}_t)$. $(\widehat{c}_t)$ is dedicated to encoding the bias induced by the mode of variation in the current sequence. To control which information constitutes a bias of the variation and should be added to the memory cell, an input gate $(\widehat{i}_t)$ and a forget gate $(\widehat{f}_t)$ are also added. To ensure that the introduced cell encodes the bias of the variation, a static sequence $(\widehat{x}_{\tau})$ is fed to the mode variational LSTM. The static sequence $(\widehat{x}_{\tau})$ is obtained by replicating a frame sampled from the input sequence $(x_{t})$ at a certain time $\tau$. Note that in  the experiments (section 4) of this paper, $\tau$ was set to 0 (beginning of video sequence which has neutral expression), but $\tau$ could be set arbitrarily for encoding the bias of the mode variation (please see the experiment in section 4.4). The bottom part of the mode variational LSTM, shown in green in Figure~\ref{fig2:b}, shows the static sequence path for encoding the bias of the variation at a certain time $\tau$. Accordingly, the bias of the variation is encoded using:

\begin{equation} 
	\label{eq2}
	\begin{split}
		&{\hat{i}}_t= \sigma{}(W_{\hat{x}\hat{i}}{\hat{x}}_{\tau{}}+W_{\hat{h}\hat{i}}{\hat{h}}_{t-1}+W_{\hat{c}\hat{i}}{\hat{c}}_{t-1}+b_{\hat{i}}),\\
		&{\hat{f}}_t= \sigma{}(W_{\hat{x}\hat{f}}{\hat{x}}_{\tau{}}+W_{\hat{h}\hat{f}}{\hat{h}}_{t-1}+W_{\hat{c}\hat{f}}{\hat{c}}_{t-1}+b_{\hat{f}}),\\
		&{\hat{c}}_t=f_t{\hat{c}}_{t-1}+i_t tanh\left(W_{\hat{x}\hat{c}}{\hat{x}}_{\tau{}}+W_{\hat{h}\hat{c}}{\hat{h}}_{t-1}+b_{\hat{c}}\right),\\
		&{\hat{h}}_t=o_t tanh({\hat{c}}_t).\\
	\end{split}
\end{equation}

To encode the dynamics important for the task at hand (e.g., FER), the dynamic sequence is fed to the mode variational LSTM. In retrospect, the encoding of the dynamics of the sequence in the mode variational LSTM are as follows: 

\begin{equation} 
	\label{eq3}
	\begin{split}
		&i_t= \sigma{}(W_{xi}x_t+W_{hi}h_{t-1}+W_{ci}c_{t-1}+b_i),\\
		&f_t= \sigma{}(W_{xf}x_t+W_{hf}h_{t-1}+W_{cf}c_{t-1}+b_f),\\
		&c_t=f_tc_{t-1}+i_t tanh(W_{xc}x_t+W_{hc}h_{t-1}+b_c),\\
		&h_t=o_t tanh\left(c_t\right).\\
	\end{split}
\end{equation}

Finally, to obtain the latent features from both cells (i.e., the cell responsible for encoding the dynamics and the cell responsible for encoding the bias) with respect to the changes of the input frames, the output gate is shared between both cells as:

\begin{equation} 
	\label{eq4}
	\begin{split}
		o_t= &\sigma{}(W_{xo}x_t+W_{ho}h_{t-1}+W_{co}c_t+ \\
	 		 &W_{\hat{c}o}{\hat{c}}_t+W_{\hat{h}o}{\hat{h}}_{t-1}+b_o).\\
	\end{split}
\end{equation}

Notice that only one output gate is used. This is primarily the case for two reasons: (1) to synchronize the dynamics latent features $(h_t)$ with the latent features representing the bias in the variation $(\widehat{h}_t)$. (2) Incorporating the bias cell state $(\widehat{c}_t)$ and the previous latent features representing the bias in the variation $(\widehat{h}_{t-1})$, suppresses the effect of the bias on the encoded dynamics latent feature $(h_t)$.
	
	\subsection{Cross-cell peephole mode variational LSTM}
	\begin{figure*}[!ht]

	\begin{minipage}[c][6.5cm][t]{0.5\textwidth}
		\vspace*{\fill}
		\centering
			\includegraphics[width=0.85\textwidth,keepaspectratio]{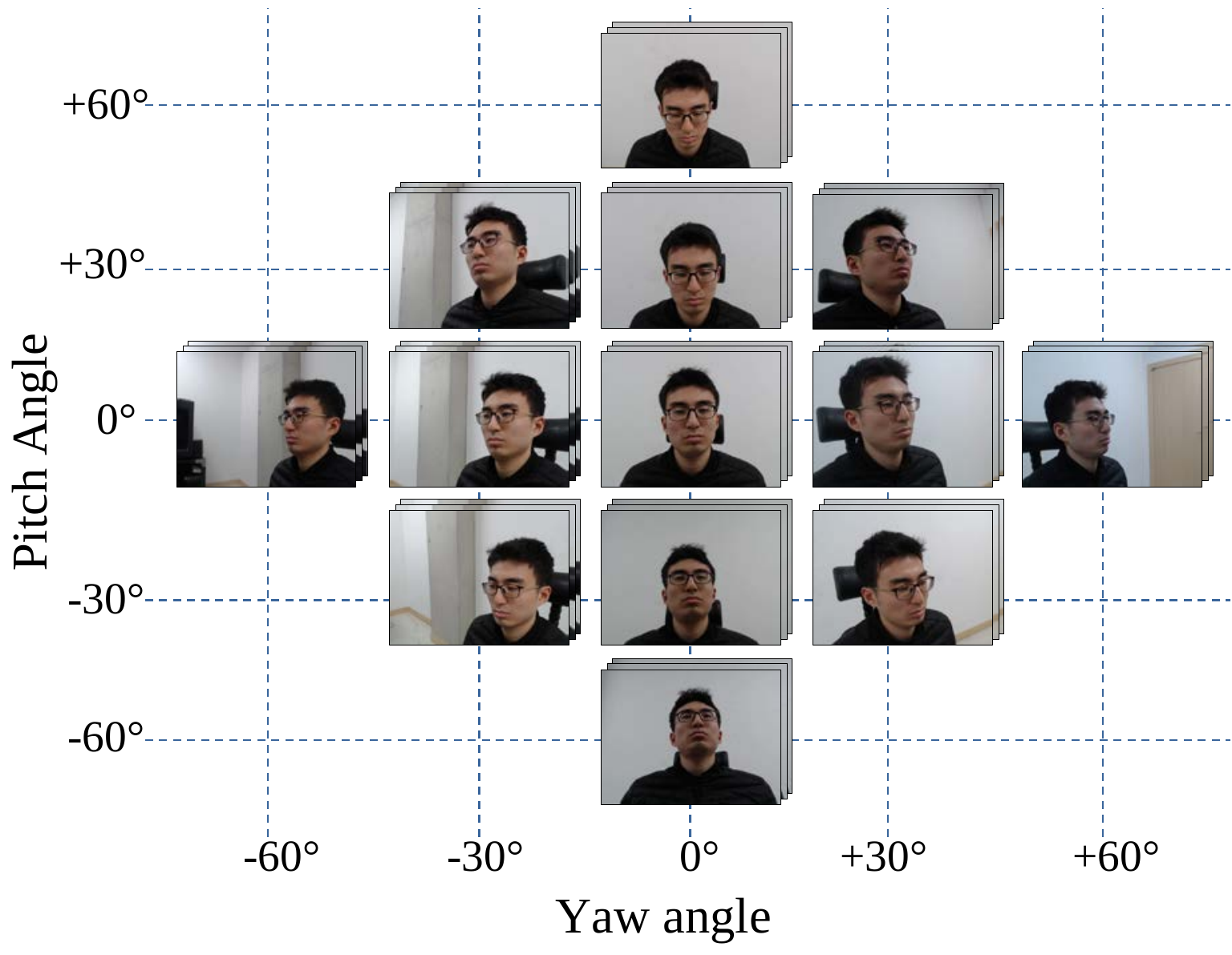}
		\subcaption{}
		\label{fig3:a}
	\end{minipage}
	\begin{minipage}[c][6cm][c]{0.5\textwidth}
		\centering
		\vspace*{\fill}
			\includegraphics[width=0.45\textwidth,keepaspectratio]{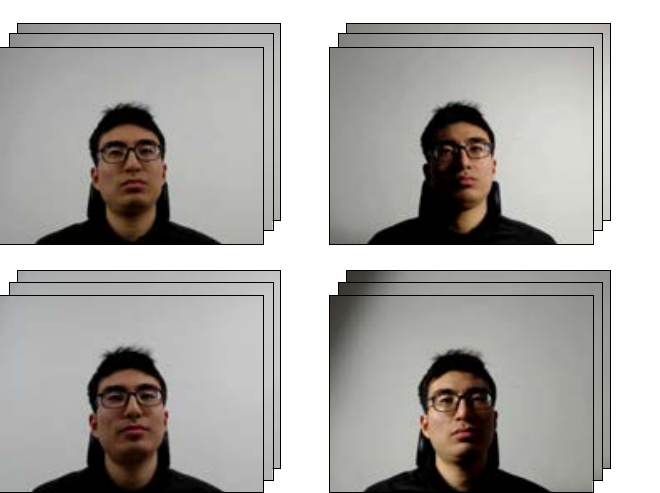}
		\subcaption{}
		\label{fig3:b}

			\includegraphics[width=0.45\textwidth,keepaspectratio]{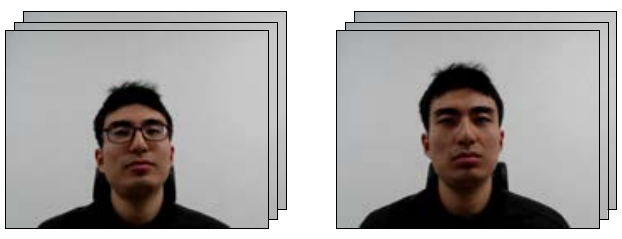}
		\subcaption{}
		\label{fig3:c}
	\end{minipage}

\caption{Examples from the KAIST Face MPMI dataset. (a) Pose variations. (b) Illumination variations. (c) Eye glasses variations.}
\label{fig:3}
\end{figure*}

		\cite{15} has reported that adding a peephole connection could improve the LSTM performance. A peephole connection mainly implies that the input gate and the forget gate can peep into the previous cell state $(c_{t-1})$), and the output gate peeps into the current cell state $(c_t)$. In other words, the cell states are utilized during the activation of the LSTM gates. The inclusion of a peephole to the LSTM has been shown to result in more discriminative features, because the LSTM gates incorporate previous states in the encoding of the new features.

Inspired by the peephole, we devise a variation of the mode variational LSTM that incorporates a peephole between the two cell states. We name it as cross-cell peephole. The cross-cell peephole can be achieved by incorporating the other cell during the encoding of the input and forget gates. The input and forget gates with cross-cell peepholes can be simply obtained by:

\begin{equation} 
	\label{eq5}
	\begin{split}
		&i_t= \sigma{}(W_{xi}x_t+W_{hi}h_{t-1}+W_{ci}c_{t-1}+W_{\hat{c}i}{\hat{c}}_{t-1}+b_i),\\
		&f_t= \sigma{}(W_{xf}x_t+W_{hf}h_{t-1}+W_{cf}c_{t-1}+W_{\hat{c}f}{\hat{c}}_{t-1}+b_f),\\
		&{\hat{i}}_t= \sigma{}(W_{\hat{x}\hat{i}}{\hat{x}}_{\tau{}}+W_{\hat{h}\hat{i}}{\hat{h}}_{t-1}+W_{\hat{c}\hat{i}}{\hat{c}}_{t-1}+ W_{ci}c_{t-1}+b_{\hat{i}}) ,\\
		&{\hat{f}}_t= \sigma{}(W_{\hat{x}\hat{f}}{\hat{x}}_{\tau{}}+W_{\hat{h}\hat{f}}{\hat{h}}_{t-1}+W_{\hat{c}\hat{f}}{\hat{c}}_{t-1}+ W_{cf}c_{t-1}+b_f) .\\
	\end{split}
\end{equation}

The reasoning behind employing the cross-cell peephole can be explained as follows: (1) when encoding the dynamics spatio-temporal features, it is important to know what part of the input frames constitutes a bias. By referencing the bias cell state, it would be easier for the input gate and the forget gate to encode information only important for the dynamics of the sequence. (2) It is safe to assume that not all the information in the static sequence constitute a bias of the current sequence mode of variation. Therefore, including information from the current dynamics cell-state in the input and forget gates controlling the encoding the bias of the variation would focus on removing the redundancies in the information that does not constitute a bias. This is important because the static sequence used in encoding the bias induced by the variation is obtained by replicating one frame for $N$ times. Hence, this cross-cell peephole can help retain features important to encoding the bias and dismissing irrelevant features, when the selection of the static sequence is not very clean (refer to the experiment in section 4.4).


\section{Experiments}

	\subsection{Experimental setup}
		To verify the effectiveness of the proposed mode variational LSTM, comparative experiments were conducted on the Oulu-CASIA facial expression dataset \cite{16} and the AFEW dataset \cite{R_1}. Moreover, we comprehensively evaluated the mode variational LSTM on a new dynamic facial expression dataset, which was collected under different modes of variation including pose and illumination. The dataset is named KAIST face multi-pose multi-illumination (KAIST Face MPMI) dataset. The construction of the datasets was performed as follows:

\noindent \textbf{1) Oulu-CASIA dataset \cite{14}:} Sequences of the six basic expressions (i.e., angry, disgust, fear, happy, sad, and surprise) were collected from 80 subjects under three illumination conditions. For the experiments, a total of 480 image sequences were collected from sequences captured with a visible light camera under normal illumination conditions. For each subject, the basic expression sequence was captured from a neutral face until the expression apex. 

\noindent \textbf{2) AFEW dataset \cite{R_1}:} To emulate real world conditions, the AFEW dataset is a collection of video clips collected from movies. Each sequence depicts a spontaneous expressions in an uncontrolled environment. According to the protocol defined by the EmotiW \cite{R_2}, the database is divided into three sets: training, validation, and test. Each set includes 7 different facial expressions (6 basic expressions and neutral expression). As the ground truth of test set is unreleased, the results are compared only on the validation set. Note that, frames that the face region was not detected (a face was not available) were neglected. Only frames where the face region was automatically detected were utilized.  

\noindent \textbf{3) KAIST Face Multi-pose Multi-illumination dataset (KAIST Face MPMI):} Sequences of the seven expressions (6 basic expressions and a neutral expression sequence) were collected from 104 subjects. Each expression sequence was recorded via 13 web cameras simultaneously, resulting in 13 pose variations. The subjects were then asked to perform the expression again under a different illumination variation. Four illumination variation conditions were recorded (room illumination condition, bright illumination condition, left illumination condition and right illumination condition). Finally, the recording was performed on two sessions, one session was obtained while the subjects were wearing eyeglasses and one without eyeglasses. Examples of the modes of variation recorded by the dataset are shown in Figure~\ref{fig:3}. 

All the experiments in this paper were conducted in a subject independent manner, such that the subjects in the training set were excluded from the test set. In particular, 10-fold cross validation \cite{17} was used for the experiments conducted on Oulu-CASIA and KAIST Face MPMI datasets. The training and validation sets of the AFEW dataset were predefined according to the protocol defined by the EmotiW \cite{R_2}. The face region was detected and facial landmark detection was performed \cite{18} on each frame. The face region was then automatically cropped and aligned based on the eye landmarks \cite{19}.

The implementation of the networks in this paper was done using TensorFlow \cite{20}. As a reference model, the network architecture used in \cite{13,21,22} was utilized. The reference model encodes the spatial features of each frame using a convolutional network (CNN). The LSTM then encodes the dynamics using the encoded spatial features as input. In this paper, the CNN initial learning rate was set to 0.0001. The training was performed for 30 epochs. For the mode variational LSTM and the LSTM \cite{13}, the learning rate was set to 0.0001. The LSTM training was conducted for 50 epochs.

To avoid overfitting, each frame in the sequence was augmented during the network training \cite{13,21,22}. 54 augmentation variations of each expressive image were obtained by: (1) horizontal flipping of the sequence frames, (2) rotating the frames between the angles $[-5^\circ , 5^\circ]$ with an increment of $1^\circ$, (3) translating the frames along $[\pm3, \pm3]$ pixels in the x and y axis with 1 pixel increments, and (4) scaling the frames with scaling factors of 0.90, 0.95, 1.05 and 1.10.

	\subsection{Effectiveness of the proposed mode variational LSTM compared to previous methods}
\begin{table}[!b]
  \centering
  \caption{Performance comparison with existing FER methods on the Oulu-CASIA dataset in terms of recognition rate.}
  \label{Table:1}
  \resizebox{\columnwidth}{!}{
    \begin{tabular}{p{6.cm}p{1.cm}}
    \toprule
    \toprule
    \multicolumn{1}{c}{Method} & \multicolumn{1}{c}{Rec. \newline{} rate(\%)} \\
    \midrule
    \midrule

    \multicolumn{1}{p{6.cm}}{LBP-TOP \cite{23}} 				& \multicolumn{1}{c}{68.13} \\
    \multicolumn{1}{p{6.cm}}{HOG 3D \cite{24}}  				& \multicolumn{1}{c}{\multirow{2}{*}{70.63}} \\
    \multicolumn{1}{p{6.cm}}{AdaLBP \cite{16}}  				& \multicolumn{1}{c}{73.54} \\
	\multicolumn{1}{p{6.cm}}{Atlases \cite{25}}  				& \multicolumn{1}{c}{75.52} \\
	\multicolumn{1}{p{6.cm}}{ExpLet \cite{26}} 				& \multicolumn{1}{c}{76.65} \\
    \multicolumn{1}{p{6.cm}}{Dis-ExpLet \cite{30}} 				& \multicolumn{1}{c}{79.00} \\
    \multicolumn{1}{p{6.cm}}{Lomo \cite{27}} 			& \multicolumn{1}{c}{82.10} \\
    \multicolumn{1}{p{6.cm}}{DTAGN \cite{17}} 					& \multicolumn{1}{c}{81.46} \\
    
	\midrule
    \multicolumn{1}{p{6.cm}}{LSTM \cite{13}}  			& \multicolumn{1}{c}{78.21} \\    
    \multicolumn{1}{p{6.cm}}{Mode variational LSTM}			& \multicolumn{1}{c}{\textbf{83.94}} \\
    \multicolumn{1}{p{6.cm}}{Mode variational LSTM \newline{} with cross-cell peephole} 		& \multicolumn{1}{c}{\multirow{2}{*}{\textbf{85.18}}} \\
    
    \bottomrule
    \bottomrule
    \end{tabular}%
}
\end{table}%

\begin{table}[!b]
  \centering
  \caption{Performance comparison with existing FER methods on the AFEW dataset in terms of recognition rate.}
  \label{Table:R_1}
  \resizebox{\columnwidth}{!}{
    \begin{tabular}{p{5.cm}p{2.0cm}p{2.0cm}}
    \toprule
    \toprule
    \multicolumn{1}{c}{\multirow{2}{*}{Method}} & \multicolumn{1}{p{1.5cm}}{Visual rec. \newline{} rate(\%)} & \multicolumn{1}{p{1.7cm}}{Multimodal \newline{} rec. rate(\%)} \\
    
    \midrule
    \midrule
    \multicolumn{1}{p{4.5cm}}{Spatio-tempral RBM \cite{R2_1}} & \multicolumn{1}{c}{\multirow{2}{*}{46.36}} & \multicolumn{1}{c}{\multirow{2}{*}{-}} \\
    \multicolumn{1}{p{4.5cm}}{Semidefinite cone \cite{R2_2}}  & \multicolumn{1}{c}{\multirow{2}{*}{39.94}} & \multicolumn{1}{c}{\multirow{2}{*}{-}} \\    
    \multicolumn{1}{p{4.5cm}}{SSE for emotion recognition \cite{RR_7}}  & \multicolumn{1}{c}{\multirow{2}{*}{46.48}} & \multicolumn{1}{c}{\multirow{2}{*}{59.01}} \\
    \multicolumn{1}{p{4.5cm}}{Temporal multimodal fusion \cite{RR_8}}  & \multicolumn{1}{c}{\multirow{2}{*}{48.60}} & \multicolumn{1}{c}{\multirow{2}{*}{52.20}} \\
    \multicolumn{1}{p{4.5cm}}{HoloNet \cite{RR_9}}  & \multicolumn{1}{c}{44.57} & \multicolumn{1}{c}{51.96} \\    
    \multicolumn{1}{p{4.5cm}}{AU-Aware facial features\cite{RR_10}}  & \multicolumn{1}{c}{\multirow{2}{*}{45.39}} & \multicolumn{1}{c}{\multirow{2}{*}{49.09}} \\	
    \multicolumn{1}{p{4.5cm}}{RNN \cite{R1_1}}  & \multicolumn{1}{c}{\multirow{2}{*}{39.6}} & \multicolumn{1}{c}{\multirow{2}{*}{54.716}} \\
		
	\midrule
	\multicolumn{1}{p{4.5cm}}{Mode variational LSTM} 	& \multicolumn{1}{c}{\textbf{48.83}} & \multicolumn{1}{c}{-} \\
	
	\multicolumn{1}{p{4.5cm}}{Mode variational LSTM \newline{} with cross-cell peephole} 	& \multicolumn{1}{c}{\multirow{2}{*}{\textbf{51.44}}} & \multicolumn{1}{c}{\multirow{2}{*}{-}} \\
    \midrule
    \midrule
  
    \end{tabular}%
    }
\end{table}%

\begin{table}[!b]
  \centering
  \caption{Performance comparison between an LSTM and the proposed mode variational LSTM on the KAIST Face MPMI dataset in terms of recognition rate.}
  \label{Table:2}
  \resizebox{\columnwidth}{!}{
    \begin{tabular}{p{5.cm}p{1.8cm}}
    \toprule
    \toprule
    \multicolumn{1}{c}{Method} & \multicolumn{1}{c}{Recognition rate(\%)} \\
    \midrule
    \midrule
  
    \multicolumn{1}{p{5.cm}}{LSTM \cite{13}}  			& \multicolumn{1}{c}{79.93} \\    
    \multicolumn{1}{p{5.cm}}{Mode variational LSTM}			& \multicolumn{1}{c}{\textbf{82.61}} \\
    \multicolumn{1}{p{5.cm}}{Mode variational LSTM \newline{} with cross-cell peephole} 		& \multicolumn{1}{c}{\multirow{2}{*}{\textbf{84.98}}} \\
    
    \bottomrule
    \bottomrule
    \end{tabular}%
}
\end{table}%

		To demonstrate the effectiveness of the proposed mode variational LSTM, the FER performance of the proposed method was compared to previously reported state-of-the-art and existing methods on the Oulu-CASIA dataset. The experiment was conducted under 10-fold subject-independent cross validation and the prediction \cite{13,17}. The comparative recognition rates are shown in Table~\ref{Table:1}. As shown in the table, the proposed method outperformed existing state-of-the-art FER methods. Specifically, the proposed method showed better recognition rates compared to the deep learning based methods with spatio-temporal features. The proposed method out-performs methods utilizing RNNs and LSTM. This is attributed to the efficient encoding of the expression dynamics and the robustness of the proposed mode variational LSTM towards modes of variation in the test data. 

To further evaluate the effectiveness of the proposed mode variational LSTM to modes of variation, we performed a comparative experiment on the AFEW dataset and the KAIST Face MPMI dataset. Both datasets contain a large number of variations. For the AFEW dataset, Table~\ref{Table:R_1} shows a comparison with previously recorded methods(including spatio-temporal and appearance based methods) \cite{R2_1,R2_2,RR_7,RR_8,RR_9,RR_10,R1_1}. On the other hand, no previously recorded methods are available on the KAIST Face MPMI datasets. As a baseline performance, the method in \cite{13} has been utilized. For a fair comparison, the CNN part of the method in \cite{13} was used to obtain the spatial features. Then, the comparison was performed between the LSTM (as described in \cite{15}) and the proposed mode variational LSTM. The comparative results are shown in Table~\ref{Table:2}. As seen from the results in Table~\ref{Table:R_1} and Table~\ref{Table:2}, the proposed method significantly improves the FER performance on the both datasets. These results verify that the proposed method generates more discriminative features regardless of the mode variations.

	\subsection{Effectiveness of the proposed mode variational LSTM in encoding spatio-temporal features robust to unseen mode variations}
		\begin{figure}
    \centering
    
    \begin{subfigure}[b]{1\linewidth}
                 \centering
                 \includegraphics[width=0.8\linewidth,keepaspectratio]{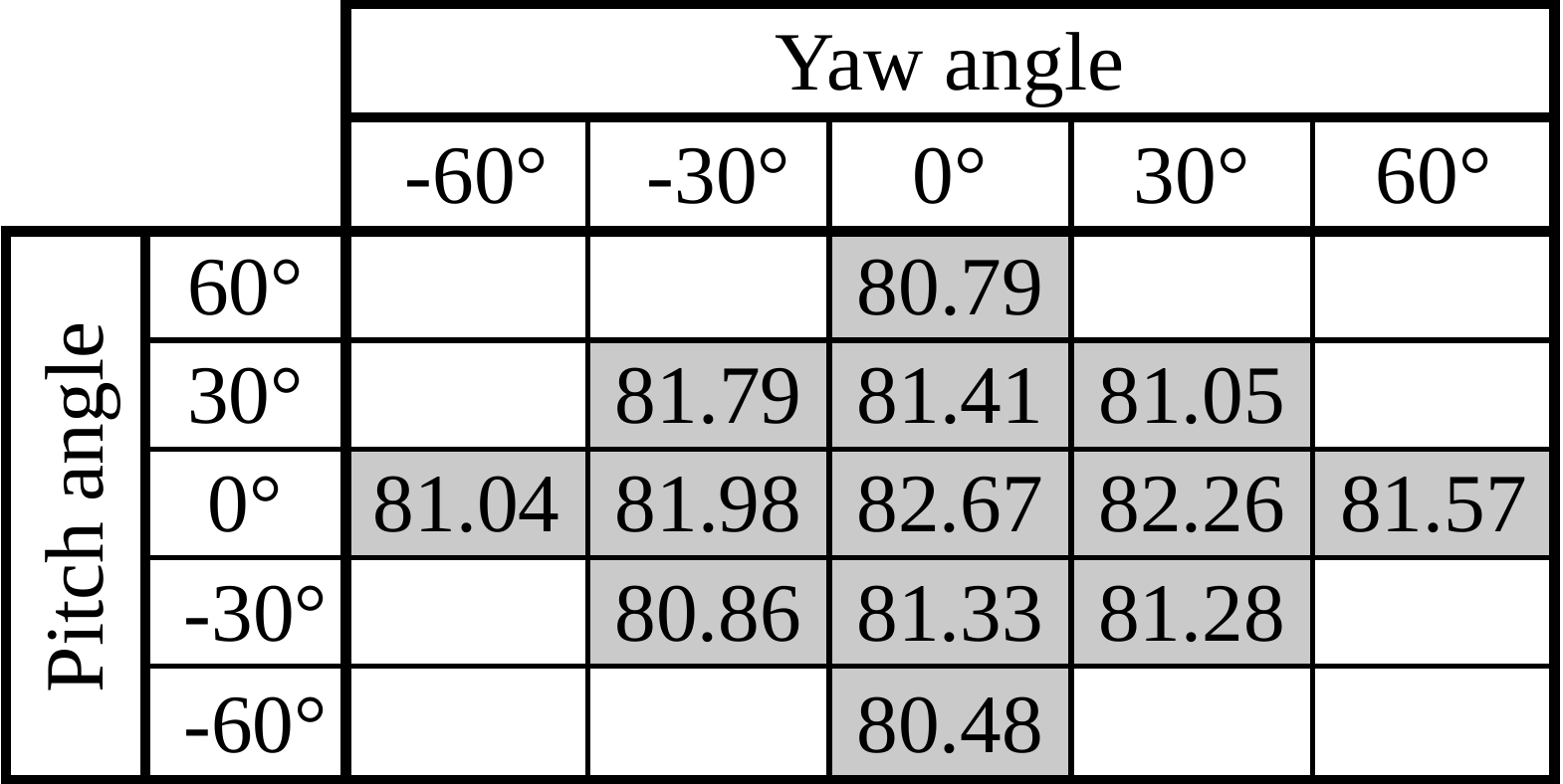}
                 \caption{}
             	 \label{fig4:a}
         \end{subfigure}
         

     \begin{subfigure}[b]{1\linewidth}
             \centering
             \includegraphics[width=0.8\textwidth,keepaspectratio]{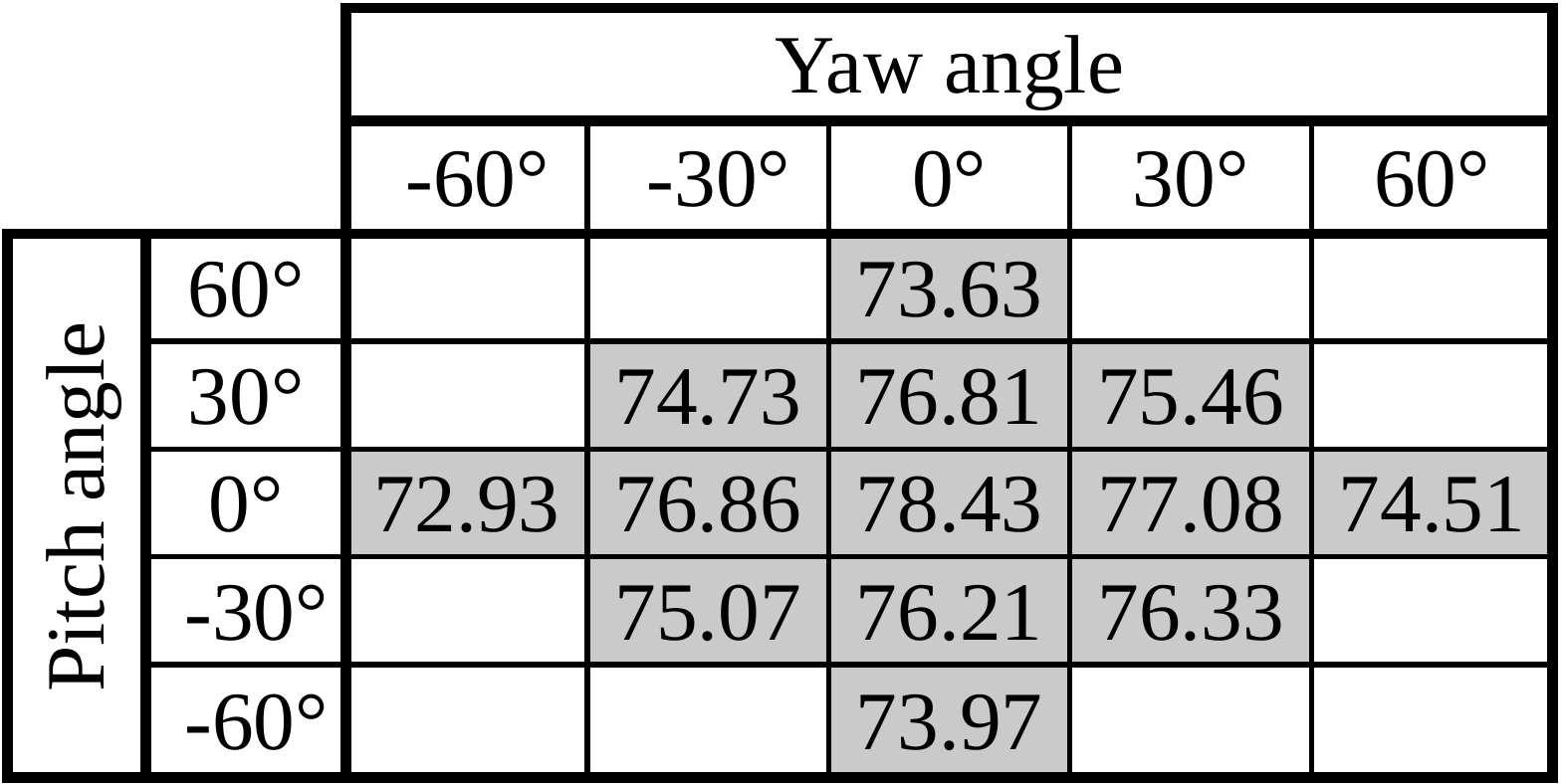}
             \caption{}
             \label{fig4:b}
     \end{subfigure}

   \caption{Performance comparison on unseen pose variations in terms the of recognition rate (\%). (a) The proposed mode variational LSTM with cross-cell peephole. (b) LSTM \cite{13}.}
\label{fig:4}
\end{figure}
	
		In this experiment, we investigate the effectiveness of the proposed mode variational LSTM in encoding spatio-temporal features robust to unseen mode variations. To that end, three types of unseen mode variations (i.e., appearance variations, pose illumination variations and pose variations) were used to compare the proposed mode variational LSTM with the LSTM \cite{13}. First, to evaluate the robustness of the proposed mode variational LSTM towards appearance variations, a cross-racial evaluation experiment on the Oulu-CASIA dataset was performed. The Oulu-CASIA dataset is divided into Asian and Finnish (Caucasian) subjects, collected at two different sessions. It is known that the facial structure (appearance) can play a crucial role in the appearance of the presented facial expression \cite{28,29}. Hence, when training with an exclusive subset of the dataset with a single race, the FER model performance on the unseen appearance is expected to be suboptimal. Table~\ref{Table:3} shows the obtained recognition rate when the proposed method was trained exclusively on the Asian subjects and validated on the Caucasian subjects and vice versa. For comparison, a LSTM \cite{13,15} was trained to encode the dynamics of the facial expression with the same condition. The results show that the proposed method improves the performance of the FER in terms of the recognition rate.

In the other two parts of this experiment, we evaluate robustness of the proposed mode variational LSTM towards the illumination and pose variations. To that end, we resolved to the KAIST Face MPMI dataset. To evaluate the robustness of the proposed method towards illumination variations, we trained the mode variational LSTM model and the LSTM model using only the sequences captured in the room illumination conditions. During the test phase, sequences from all the illumination variations were used (i.e., sequences with room illumination condition, bright illumination condition, left illumination condition and right illumination condition). The experiments were done in a subject independent 10-fold cross validation. The recognition rate on each illumination variation was obtained and shown in Table~\ref{Table:4}.  Not only do the results show a more consistent recognition performance on unseen illumination variations, but improve the recognition performance on the same illumination variation. This can be attributed to the fact that the bias induced by the subject appearance is also reduced.

Finally, we validate the robustness of the proposed mode variational LSTM toward pose variations. We evaluate the FER performance on unseen pose variations. To that end, we trained the mode variational LSTM model and the LSTM model \cite{13} using only the frontal facing sequences (i.e., yaw angle and pitch angles are set to zero). During the test phase, sequences from all pose variations were used. The experiments were done in a subject independent 10-fold cross validation. The recognition rate on each pose variation was obtained and shown in Figure~\ref{fig:4}.  As seen from the figure, the proposed mode variational LSTM sustains a high recognition rate over all the poses. On the other hand, a steep degradation in the FER performance on other pose variations was obtained using the LSTM.

\begin{table}
  \centering
  \caption{Cross-race FER performance comparison with the LSTM in terms of recognition rate (\%).}
  \label{Table:3}
  \resizebox{\columnwidth}{!}{
    \begin{tabular}{p{3.4cm}p{1cm}p{1.7cm}p{1.7cm}}
    \toprule
    \toprule
	
	\multicolumn{2}{c}{ }		& \multicolumn{2}{p{3.4cm}}{Cross-race evaluation \newline{} test set}\\\cmidrule{3-4}
	\multicolumn{1}{c}{Method}  & \multicolumn{1}{c}{Training Set}	 	& \multicolumn{1}{p{1.3cm}}{Asian}  & \multicolumn{1}{p{1.3cm}}{Caucasian} \\										
		
    \midrule
    \midrule
	
	\multirow{2}{*}{LSTM \cite{13}} 	& \multicolumn{1}{p{1.cm}}{Asian}	    & \multicolumn{1}{p{1.3cm}}{-}		& \multicolumn{1}{p{1.3cm}}{73.39}\\
																& \multicolumn{1}{p{1.cm}}{Caucasian} & \multicolumn{1}{p{1.3cm}}{74.67}	& \multicolumn{1}{p{1.3cm}}{-}\\	
	\midrule
	
	\multicolumn{1}{p{3.0cm}}{\multirow{2}{4.0cm}{Mode variational LSTM \newline{} with cross-cell peephole}} 	& \multicolumn{1}{p{1.cm}}{Asian} 	& \multicolumn{1}{p{1.3cm}}{-}		& \multicolumn{1}{p{1.3cm}}{\textbf{79.14}}\\
																& \multicolumn{1}{p{1.cm}}{Caucasian} & \multicolumn{1}{p{1.3cm}}{\textbf{80.95}}	& \multicolumn{1}{p{1.3cm}}{-}\\

    \bottomrule
    \bottomrule
    \end{tabular}%
}
\end{table}%

\begin{table}
  \centering
  \caption{FER performance comparison with LSTM on unseen illumination variations in terms of recognition rate (\%).}
  \label{Table:4}
  \resizebox{\columnwidth}{!}{
    \begin{tabular}{p{3.4cm}p{1.0cm}p{1.0cm}p{1.0cm}p{1.0cm}}
    \toprule
    \toprule
    \multicolumn{1}{c}{ }					& \multicolumn{4}{c}{Illumination condition}\\ \cmidrule{2-5}
    \multicolumn{1}{c}{Method} & \multicolumn{1}{c}{Room \newline{} condition}	& \multicolumn{1}{c}{bright}	& \multicolumn{1}{c}{Left} & \multicolumn{1}{c}{Right}\\
    \midrule
    \midrule
  
    \multicolumn{1}{p{3.4cm}}{LSTM \cite{13}}  	& \multicolumn{1}{c}{78.21} & \multicolumn{1}{c}{76.98} & \multicolumn{1}{c}{73.67} & \multicolumn{1}{c}{74.39} \\    
    \multicolumn{1}{p{3.4cm}}{Mode variational LSTM with cross-cell peephole}	& \multicolumn{1}{c}{\multirow{2}{*}{\textbf{83.11}}}	& \multicolumn{1}{c}{\multirow{2}{*}{\textbf{83.03}}} & \multicolumn{1}{c}{\multirow{2}{*}{\textbf{82.87}}} & \multicolumn{1}{c}{\multirow{2}{*}{\textbf{82.91}}} \\

    \bottomrule
    \bottomrule
    \end{tabular}%
}
\end{table}%

	\subsection{Effectiveness of the cross-cell peephole in retaining useful bias information}
\begin{table}[!ht]
  \centering
  \caption{FER performance comparison in terms of recognition rate (\%) between mode variational LSTM with and without the cross-cell peephole for randomly selected $\tau$. Note that, the static sequence $(\widehat{x}_{\tau})$ was generated by replicating the frame at the randomly selected $\tau$.}
  \label{Table:5}
  \resizebox{\columnwidth}{!}{
    \begin{tabular}{p{3.4cm}p{1.8cm}p{2.cm}}
    \toprule
    \toprule
    \multicolumn{1}{c}{ }					& \multicolumn{2}{c}{Test set}\\ \cmidrule{2-3}
    \multicolumn{1}{c}{Method} & \multicolumn{1}{p{2.0cm}}{Oulu-CASIA} & \multicolumn{1}{c}{KAIST Face MPMI} \\
    \midrule
    \midrule
  
    \multicolumn{1}{p{3.4cm}}{Mode variational LSTM}			& \multicolumn{1}{c}{81.43} & \multicolumn{1}{c}{80.91}\\
    \multicolumn{1}{p{3.4cm}}{Mode variational LSTM \newline{} with cross-cell peephole} 		& \multicolumn{1}{c}{\multirow{2}{*}{\textbf{83.57}}} & \multicolumn{1}{c}{\multirow{2}{*}{\textbf{82.67}}} \\
    
    \bottomrule
    \bottomrule
    \end{tabular}%
}
\end{table}%

		In this experiment, we show that the cross-cell peephole is beneficial in dynamically updating the bias cell state. This dynamic update improves what information should be retained as bias, and what information should be neglected with respect to the dynamic input sequence. To evaluate that, when generating the static sequence, a random frame from the dynamic sequence was used and replicated instead of replicating the first frame of the sequence (i.e., $\tau$ in eq.2 was selected randomly). This means that the static sequence is no longer guaranteed as a static sequence of neutral expression. Experiments were conducted on the Oulu-CASIA, and KAIST Face MPMI datasets. The FER performance was obtained from the mode variational LSTM with and without the cross-cell peephole. The results on both datasets are shown in Table~\ref{Table:5}. Compared to neutral frame selection, degradation in the performance of the recognition rate could be seen in the case of random frame selection when obtaining the static sequence. This can be attributed to the fact that the bias cell state retains redundant information about the facial expression along with the bias information. On the other hand, the cross-cell peephole continuously updates the cell state with respect to the dynamics sequence. As a result, it could neglect redundant information and retain information that could be considered as bias induced by the mode variation. The results seen in Table 5 show better performance than the conventional LSTM (79.93\% on the KAIST Face MPMI dataset and 78.21\% on the Oulu-CASIA dataset).

\section{Conclusion}
	In this paper, we addressed the effect of mode variations on the encoded spatio-temporal features using LSTMs. Using static sequences, we showed that the LSTM encoded spatio-temporal features retain a bias caused by modes of variation (such as illumination variations, pose variations and appearance variations). To reduce the effect of this bias on the encoded spatio-temporal features, mode variational LSTM was proposed. The mode variational LSTM modifies the original LSTM structure by adding an additional cell state that focuses on encoding the mode variation in the input video sequence. The retention of information in the bias cell state was regulated via additional gating functionality. The bias from the mode of variation was extracted from a static sequence generated by replicating a frame of the input sequence. The effectiveness of the proposed mode variational LSTM was evaluated on multiple datasets. The results showed that the proposed mode variational LSTM outperforms previous methods. Comprehensive experiments also showed that the spatio-temporal features encoded by the proposed mode variational LSTM are more robust to modes of variation unseen during the training.
	
\section{Acknowledgment}
	The authors would like to express their gratitude to Kihyun Kim, Minho Park and Seong Tae Kim for their discussion and efforts in the recording and collection of the KAIST face MPMI dataset. This work was partially supported by the Institute for Information \& communications Technology Promotion(IITP) grant funded by the Korea government(MSIT) (No. 2017-0-01778, Development of Explainable Human-level Deep Machine Learning Inference Framework). This work was also partially supported by the Institute for Information \& communications Technology Promotion (IITP) grant funded by the Korea government(MSIT) (No.2017-0-00780, Development of VR sickness reduction technique for enhanced sensitivity broadcasting).

\bibliography{refs}
\bibliographystyle{aaai}
		
\end{document}